**Nonlinearity Enhanced Adaptive Activation Functions**

David Yevick
Department of Physics and Astronomy
University of Waterloo
Waterloo ON, N2L 3G1 Canada
yevick@uwaterloo.ca

**Abstract:** A general procedure for introducing parametric, learned, nonlinearity into activation functions is found to enhance the accuracy of representative neural networks without requiring significant additional computational resources. Examples are given based on the standard rectified linear unit (ReLU) as well as several other frequently employed activation functions. The associated accuracy improvement is quantified both in the context of the MNIST digit data set and a convolutional neural network (CNN) benchmark example.

**Keywords:** Activation functions, Machine Learning, Neural Networks, Pattern Recognition

## 1. Introduction:

While neural networks (NN) were first proposed in 1943[1], initial implementations were restricted to small numbers of neurons and layers[2, 3]. These limitations were later eliminated by the backpropagation training algorithm[3–5] together with improved hardware performance. The resulting algorithm generates a system model from experimental or simulated data and can accordingly be applied to a wide variety of problems that do not possess analytic solutions. Rather than characterizing these systems by a few deterministic equations, numerous parametrized nonlinearly interacting variables are employed. By minimizing a loss function (which may be further subject to physical constraints in physics-informed machine learning[6]) these parameters are optimized. The trained model can then predict the response of the system to unobserved input data. However, while such procedures are general, they obviously lack the precision and efficiency associated with analytic methods. Additionally, the large dimensionality of the model obscures the underlying physics, which however can in certain cases be recovered by projecting onto an appropriately chosen low-dimensional space.[7] For complex systems that cannot be realistically described analytically, however, especially in the presence of stochastic noise or measurement inaccuracy, neural networks can replicate the observed behavior and hence be effectively optimal.[8–12]

Backpropagation based neural networks insert data records into the first of a series of layers of artificial neurons. Each neuron processes either the input data or the outputs of the previous layer by applying a nonlinear 'activation' function to the sum of a bias term and the product of each of its input values with a separate weighting factor. The collection of biases and weights, which is typically extensive for deep neural networks with multiple layers, is then adjusted to minimize a certain cost function.[13] In this process, the activation function nonlinearity introduces curvature into the decision boundaries, as can be visualized in PCA space,[14] significantly enhancing the discrimination capability of the network.

**2. Activation Functions:** The earliest activation function the threshold logic unit (linear threshold unit), was based on the response of biological neurons and consequently possesses the form $f(x) = \theta(x)$ in which $\theta(x)$ is the Heaviside step function.[1] However, researchers subsequently recognized that just as the optimal neural network structure is problem dependent, the activation function properties should similarly be adapted to the implementation details. Since the same activation function is applied to all the outputs of a given layer, selecting the activation function appropriately can noticeably reduce the number of neurons or layers required to achieve a desired level of accuracy. However, the improvement is somewhat limited since modifications to the activation function are partially offset by compensating variations in the network parameters during training.

Activation functions can be classified as either non-adaptive, which are invariant during training, or adaptive. Comparative studies of non-adaptive activations generally favor the rectified linear unit **relu**, which corresponds to an ideal switch and is consequently defined by[15, 16]

$$f(x) = x\Theta(x) \tag{1}$$

because of its obvious simplicity and efficiency. The nonlinearity associated with Eq.(1) originates from the slope discontinuity at $x = 0$. Although $x$ appears in Eq.(1) as a scalar variable, in the $l{:}th$ layer of a neural network it represents the vector argument $\mathbf{W}^{(l)}\vec{x}^{(l-1)} + \vec{b}^{(l)}$ of values generated from the outputs $\vec{x}^{(l-1)}$ of the previous layer, where $\boldsymbol{W}^{(l)}$ and $\vec{b}^{(l)}$ are the associated weight matrix and bias vector, respectively. The function $f$ is then applied to each component of its argument.

An improved but less numerically efficient alternative to Eq.(1) is provided by the **swish** activation function.[17]

$$f(x, \beta) = \frac{x}{1 + e^{-\beta x}} \tag{2}$$

Eq.(3) is similar in form to the **relu** but possesses a continuous derivative at the origin, which facilitates optimization. The softsign activation function replaces the exponential function in the denominator of the swish function by $|x|$ in order to decrease computation times[18] The exponential linear unit (ELU) [19]

$$f(x, \beta) = x\theta(-x) + \beta(e^{x} - 1)\theta(x) \tag{3}$$

provides a further alternative that can often prove more accurate and efficient than the ReLU.[19, 20]. Although its derivative is discontinuous at the origin, for $x < 0$ the gradient is finite, eliminating the problem of dead neurons. The function itself is negative for negative $x$ so that the average output of the neuron can be closer to zero, which reduces issues arising from vanishing or exploding gradients.[20] Numerous additional activation functions based on modifications of the ReLU as well as more involved profiles have been proposed.[20] However, these often prove less computationally efficient[15–17] although recent developments such as the Mish activation function, which in a somewhat similar fashion to the strategy of this paper employs the softplus function as an argument to the tanh activation function to achieve enhanced performance,[21] as well as the modified $\sin^2 x$ snake activation function[22] that is effective for periodic data and the GELU function,[23] which stochastically eliminates inputs depending on their values in place of a deterministic function, have led to significant improvements for different classes of problems.

Adaptive activation functions possess one or more parameters that are optimized during training. An adaptive version of Eq.(1), the parametric rectified linear unit (PReLU), is given by

$$f(x,\beta) = \beta x\theta(-x) + x\theta(x) \tag{4}$$

where a different $\beta$ value is learned in each network layer. Similarly, in the parametric ELU (PELU),[24], $\beta$ in Eq.(3) is optimized for each layer separately. However, the PReLU like the ReLU can induce exploding gradients when employed in certain deep neural network architectures. Note that the form of the ELU and PELU functions results in a fixed ratio of the magnitude of the linear and nonlinear components while the exponential function evaluation requires additional computation time. These drawbacks are shared by many other parametric activation functions.

Since the network weights and biases adjust during training to the features of the activation function, the improvement in predictive accuracy afforded by parametric functions is often not sufficient to justify the additional coding and computational effort.[17, 20, 25–31] A comprehensive analysis of activation functions is beyond the scope of this paper but can be found in e.g. [18, 20, 25, 32–36] The approach of this paper, however differs from standard methods in that it operates on classes of activation functions rather than simply presenting an individual function.

**3. Adaptive procedure:** This paper introduces a, to our knowledge, novel technique for converting nonparametric into parametric activation functions that is both simply implemented and computationally efficient. In particular, an odd binomial term with layer-dependent coefficients is incorporated into an existing activation function $f$ according to

$$f^{(l)}\left(x, c_0^{(l)}, c_1^{(l)}\right) = f\left(x\left(\, c_0^{(l)} + \gamma c_1^{(l)}|x|\right)\right) \tag{5}$$

with $\gamma$ a user-specified global constant. The amplitudes of the two terms in the activation function are employed as additional parameters at each layer $l$ of the neural network during training. While this adds $2(N-1)$ optimization parameters in the case of a $N$ layer neural network with a **softmax** or equivalent output layer, the odd quadratic term imparts the smallest possible change to the computation time and the overall activation function structure. The global constant, $\gamma$, further regulates the amplitude of this term during the initial epochs and therefore affects the local minimum that the optimization routine eventually selects. While $\gamma$ can be eliminated if the initialization ranges of the $c_0^{(l)}$ and $c_1^{(l)}$ are appropriately specified, it is generally simpler to initialize all parameters identically and instead select $\gamma$ to insure proper convergence.

The large nonlinearity and dimensionality of a neural network obscure the dependence of the predictive accuracy of a neural network on its architecture which therefore can generally only be determined qualitatively. Within this limitation, consider Eq.(5) with $f$ given for simplicity by the ReLU activation function. In this case, the rectifying structure of the ReLU is preserved in layers with positive $c_0$ and $c_1$. Certain nodes are then accordingly eliminated from the cost function during learning, corresponding to an optimized dropout layer. Since dropout layers have been determined to improve neural network performance, the prediction accuracy is generally enhanced. However, if an excessive number of neurons are removed the “dead or dying ReLU problem” appears as the network then responds slowly to a gradient based optimizer. However, the activation functions of layers for which $c_0$ and $c_1$ are not both positive are non-zero with finite slopes within adjustable ranges of $x$. The training procedure can then more effectively optimize the number of active neurons while additionally avoiding the well-known issue of vanishing gradients.

**4. Computational Results:** The main neural network employed in this paper is a straightforward adaptation of the program in p.63-66 of [37], as summarized in the appendix. The input data consists of the benchmark MNIST collection of 70,000 labeled handwritten digits discretized on a $28 \times 28$ point grid with 256 grayscale levels. These are divided into a training set of 60,000 digits and a test set of 10,000 digits and normalized to a maximum value of unity. The neural network, consists of dense 512 and 50 neuron layers with wights and biases randomly initialized in the intervals $[0, 0.01]$ and $[0, 0.1]$, respectively. The activation functions are described by Eq.(5) where the $c_{0,1}^{(i)}$ are randomly initialized in $[0.0, 0.01]$ and $f$ is the **relu** function followed by a 10 neuron output layer with a **softmax** activation function. The network is trained with a **Nadam** optimizer and a batch size of 128 over 150 epochs. If the test accuracy remains at a value less than 0.5 (and in general close to 0.1) after 15 epochs, the calculation is prematurely terminated to conserve computer resources..

Plotting the test accuracy after every 2 epochs as a function of epoch number for 60 separate calculations yields Figure 1(a) while Figure 1(b), displays a histogram of the accuracy of the final results in the range $[0.982,0.986]$. Furthermore, in 3 of the 60 cases the final accuracy fell below $A_1 = 0.5$ after 15 epochs while 5 results, $r$, were in the interval $A_1 < r < A_2 = 0.982$. The corresponding results in Figure 2 for 60 computations with the **swish** activation function, Eq.(3), demonstrate that the analyticity of the activation function induces a smooth, quasi-Gaussian distribution of the predicted values although the mean accuracy of the **swish** and **relu** results are similar. Additionally, only a single computation was terminated with $r < A_1$ while in one case $A_1 < r < A_2$. Repeating the calculation of Figure 1 with first, the ELU and then the PReLU activation functions yielded the results of Figure 3 and Figure 4, respectively. Evidently, while the PReLU histogram is slightly more accurate than that of the ELU, the corresponding convergence rate is marginally slower, presumably because of the additional computation time required to optimize the activation parameters. Both activation functions, however, are less accurate than Eq.(5).

Employing the activation function of Eq.(5) with $\gamma = 5$, significantly improves the accuracy as evident in Figure 6 for 150 calculations. At the same time, the number of non-converged results increased considerably to 63 with $r < A_1$ and 1 with $A_1 < r < A_2$. This suggests a tradeoff between predictive accuracy and convergence probability. Indeed, decreasing $\gamma$ in Eq.(3) to 1 and to 2.5 yields the results of *Figure* 5 (a) and (b), respectively for 60 realizations with $r < A_1$ and $A_1 < r < A_2$ values of (5,4) and (6, 2), respectively. As $\gamma$ is lowered a larger percentage of calculations converge but the overall accuracy decreases towards the **relu** result. This suggests that for larger $\gamma$ the parameter space is more exhaustively sampled during the initial epochs, which enhances the probability of the algorithm converging to local minima that correspond to either misleading or highly accurate predictions. Therefore, although $\gamma$ could be optimized during training, the cost function and hence the resulting $\gamma$ value would depend on the desired output behavior. The computation time associated with Eq.(5) in our implementation is $\approx 1.6$ times that of the **relu** calculation.

The adaptive activation function of Eq. (5) for $\gamma = 1$ but with a cubic term, $x^3$ in place of $x|x|$ generates the results of Figure 7 after 180 computations. To achieve a reasonable number of converged results without manually adjusting the parameter initialization ranges to insure that the activation functions are sampled primarily at positive argument values for which the cubic term is not excessively large, the argument of Eq.(3) was replaced by $c_0^{(l)} + c_1^{(l)}x + \gamma c_2^{(l)}x^3$. The resulting accuracy is less than that of Figure 1 while the number of calculations with $r < T_1$ and $T_1 < r < T_2$ increases to 88 and 19, respectively. The loss of accuracy was found as well to be still greater if an odd quartic rather than a cubic term was instead substituted for $x|x|$. The relative advantage of the quadratic activation function argument presumably results from its larger gradient at small values which facilitates gradient based optimization. However, while not considered here because of the increased computation requirements, replacing $x|x|$ by a nonlinear term of the form $\text{sign}(x)|x|^\alpha$ where $\alpha > 1$ is a real learned parameter could in principle yield increased accuracy while still preserving the desirable features of the function $f$.

To demonstrate that the improvement in accuracy is largely architecture and problem independent, two different networks were examined. The first of these is identical to that of the previous figures, but with a single 512 neuron intermediate layer in place of the 512 and 50 neuron layers. For the parameters of Figure 1, the accuracy histogram is given by Figure 8 for the activation function of Eq.(5) with $\gamma = 1$ and $f$ replaced with the **relu** and by Figure 9 for the standard non-adaptive ELU activation function. As in the two intermediate layer case, the adaptive procedure displays noticeable improvement over the standard technique.

Finally, a more complex 4 layer CNN that can be evaluated sufficiently rapidly to enable the acquisition of an accuracy histogram was examined. The code for this example is provided at https://www.tensorflow.org/tutorials/images/classification. as part of the tensorflow package. In this case, the histogram obtained by applying the standard ELU activation function to all layers (left-hand side of Figure 10) was more accurate than that obtained with the ReLU and PReLU functions. The optimal result, shown in the right-hand side of Figure 10, was however obtained when the ELU activation function was only applied to the first of the four layers while the ReLU function was employed in the remaining three layers. If Eq. (5) with $c_{0,1}^{(i)}$ randomly initialized in the interval $[0, 1]$, the **elu** activation function and $\gamma = 0.1$ is similarly employed to the first network layer while the standard non-adaptive **relu** activation function is employed the subsequent 3 layers, the figure on the left of Figure 11 results. The histogram at the right of the figure is generated if $x^3$ is employed in the first layer activation function in place of $x|x|$. As in the MNIST calculations, Eq.(5) with a quadratic nonlinearity yields the highest accuracy and exhibits a marked improvement compared to the standard activation function results.

**5. Conclusions and Future Directions:** This paper has introduced a procedure for converting non-adaptive activation functions into an adaptive functions with similar structures but higher accuracy. The method can be implemented with minimal additional programming effort and computer resources. Accordingly, the method can be applied to a problem of interest by first identifying an optimal non-adaptive activation function by e.g. trial and error and then converting this function into its corresponding adaptive counterpart. Although the resulting performance gain is problem-dependent, for the benchmark problems considered here, the increase in accuracy compared favorably to the improvements afforded by several common alternatives to the standard ReLU activation function. As well, a certain trade-off was identified which enables more accurate solutions to be obtained during multiple runs of the program at the cost of a greater number of nonconverged results. Since these, however, can be rapidly identified and removed, they do not substantially contribute to the overall computation time.

Although the form of the activation function modification considered in this paper arose from repeated numerical experiments, since the objective was to obtain a general algorithm with low computational overhead, many possibilities were not examined. These include for example more involved parametrized analytic or piecewise continuous transformations of the activation function argument as well as, as noted above, the sign of the argument multiplied by a real power of its amplitude. Such profiles, while less computationally efficient, could be advantageous if adapted to a specific problem of interest. Further, given the extensive search space, more optimal transformations of standard activation functions could exist that yield improved accuracy, a smaller number of rejected calculations or a smoother distribution of converged values.

**Appendix:** The principal modifications of the code p.63-66 of [37] are as follows:

1)

```
def linear( x ):
   return x

 class NaiveNonlinear:
   def __init__(self, input_size):
      output_size = 2
      b_shape = (output_size,)
      b_initial_value = ( tf.random.uniform(b_shape, minval = 0, maxval = 1e-2) )
      self.b = tf.Variable(b_initial_value)

   def __call__(self, inputs):
     g = 5
     return( tf.math.maximum( tf.add( self.b[0] * inputs, g * self.b[1] * tf.sign( inputs ) * inputs * inputs ), 0  ) )

   @property
   def weights(self):
      return [ self.b ]

   def set_weights(self, my_weights):
      [ self.b ] = my_weights

def model_body( train_images, train_labels ):
      model = NaiveSequential([
         NaiveDense(input_size=new_image_length, output_size = 512, activation = linear ),
         NaiveNonlinear( intermediate_size1 ),
         NaiveDense(input_size = 512, output_size = 50, activation = linear ),
         NaiveNonlinear( intermediate_size2 ),
         NaiveDense(input_size=intermediate_size2, output_size=10, activation = tf.nn.softmax )
      ])
```

2)

```
   for loop in range( numberOfRepeats ):
      optimizer = tf.keras.optimizers.Nadam()
      model_body( train_images, train_labels )
```

**Data availability statement:** The data employed in this paper is freely available as the MNIST data set at, among other sites, mnist/ at master · cvdfoundation/mnist · GitHub.

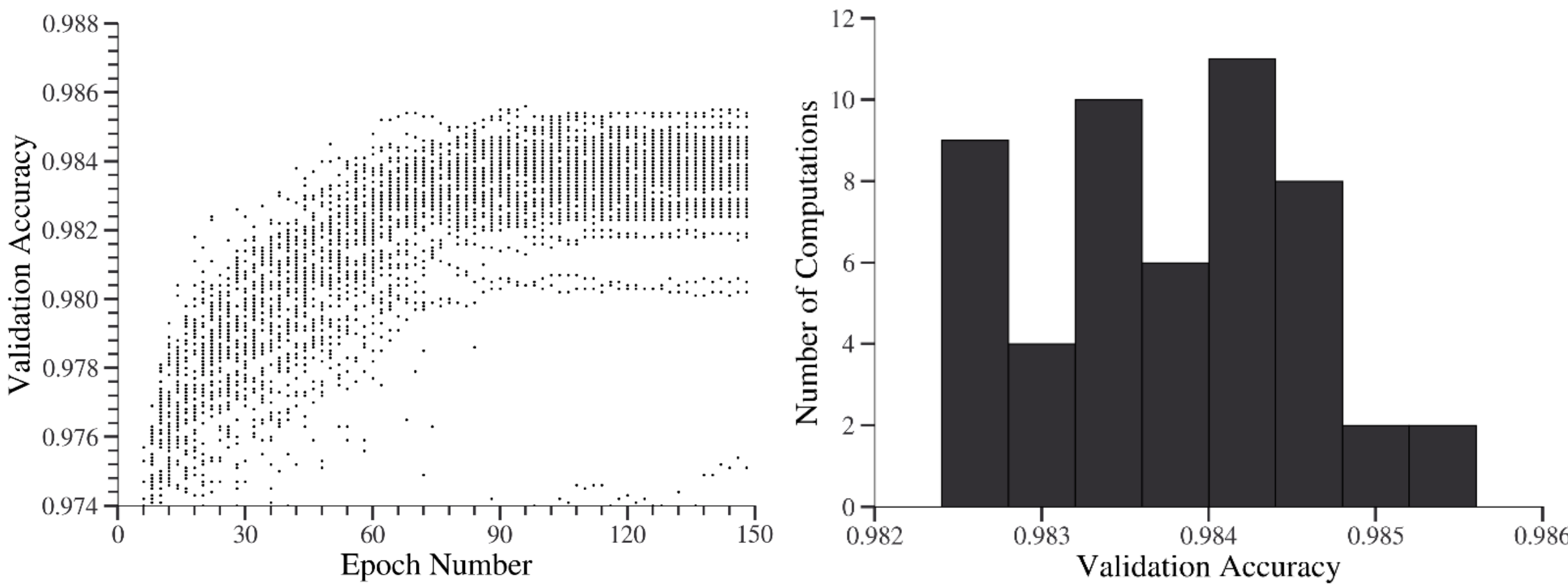

*Figure 1*: The test accuracy during and after 60 separate calculations for a 512/50/10 dense neural network with a **relu** activation function**.** In the left figure, Figure 1(a), the test accuracy is evaluated in each calculation after every 2 epochs while the right figure, Figure 1(b), displays a histogram of the final accuracy of each computation.

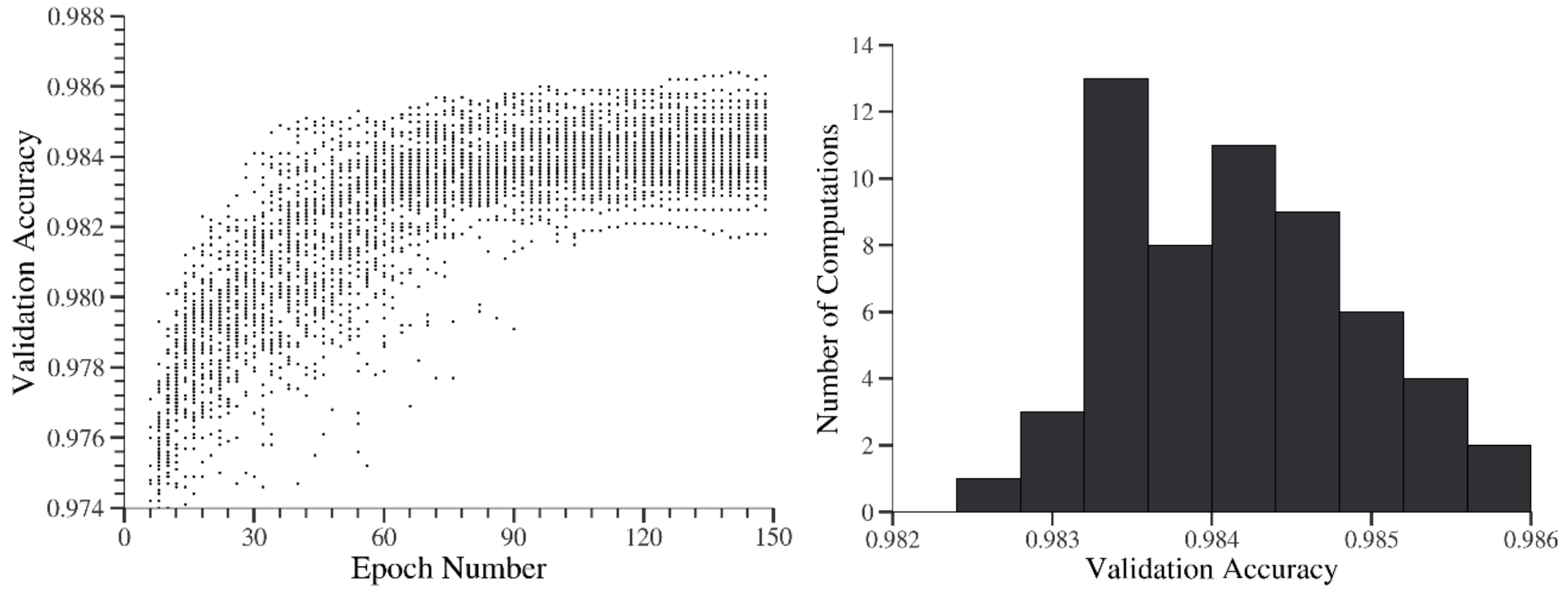

*Figure 2*: As in Figure 1 but for a ***swish*** *activation* function.

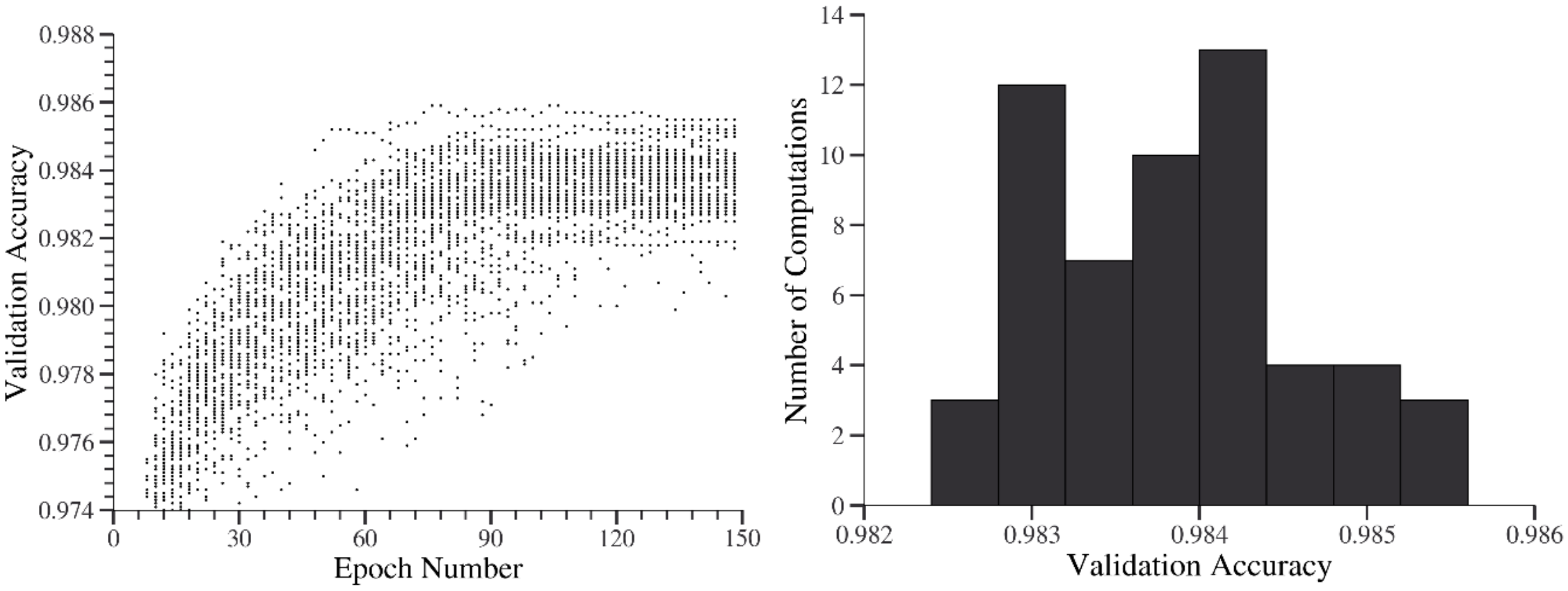

*Figure 3*: *As in Figure 1, but for the ELU activation function.*

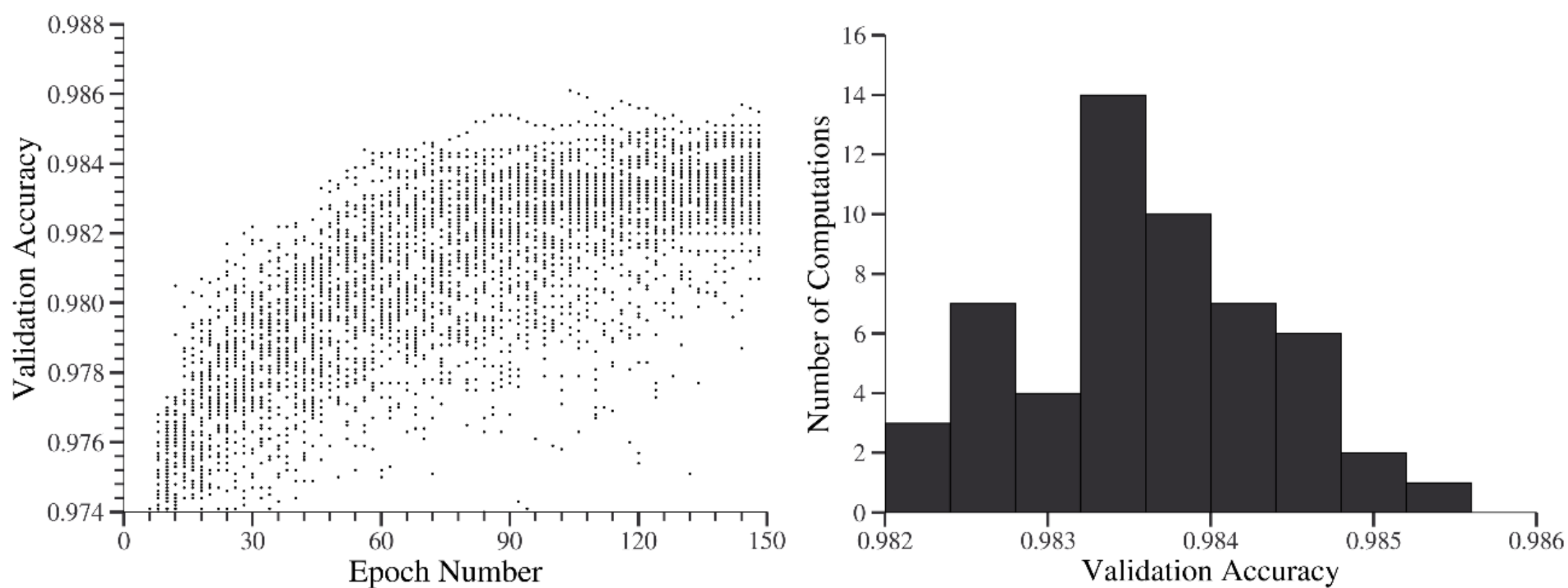

*Figure 4*: As in *Figure 1*, but for a PReLU activation function.

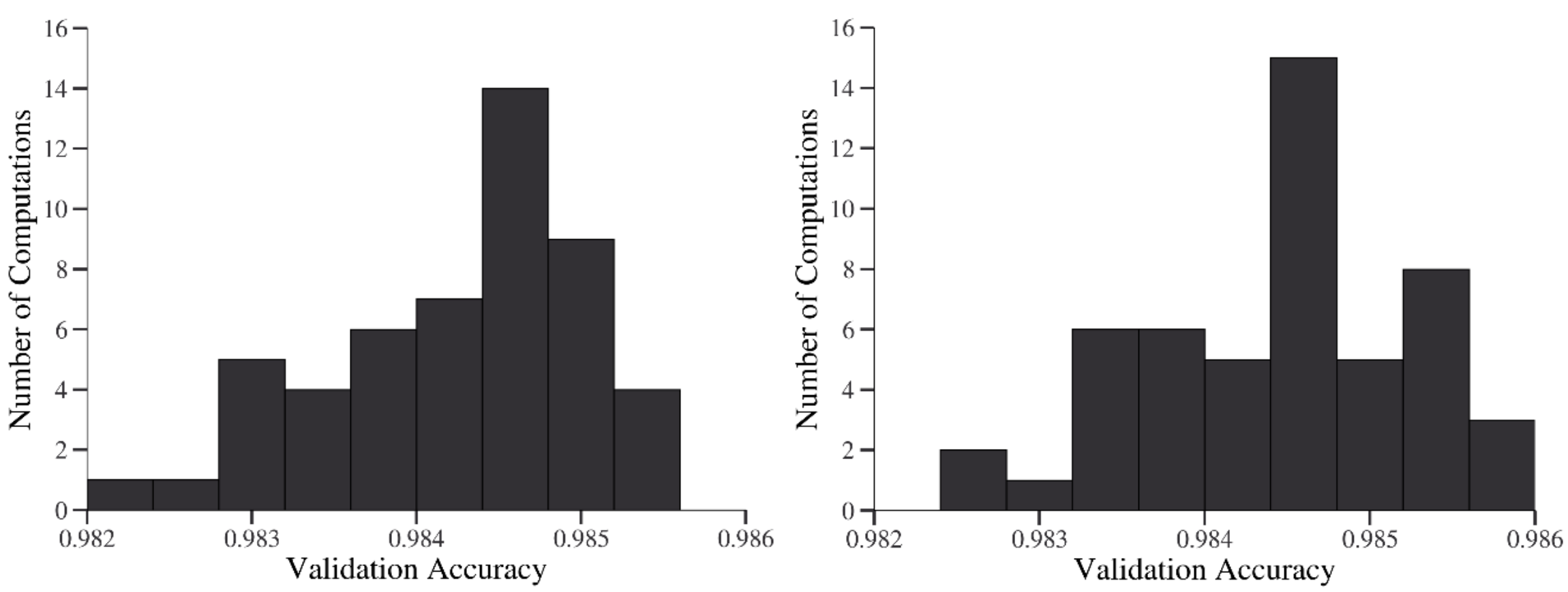

*Figure 5: The histogram of Figure 6 but for* $\gamma = 1$ *(left plot) and* $\gamma = 2.5$ *(right plot).*

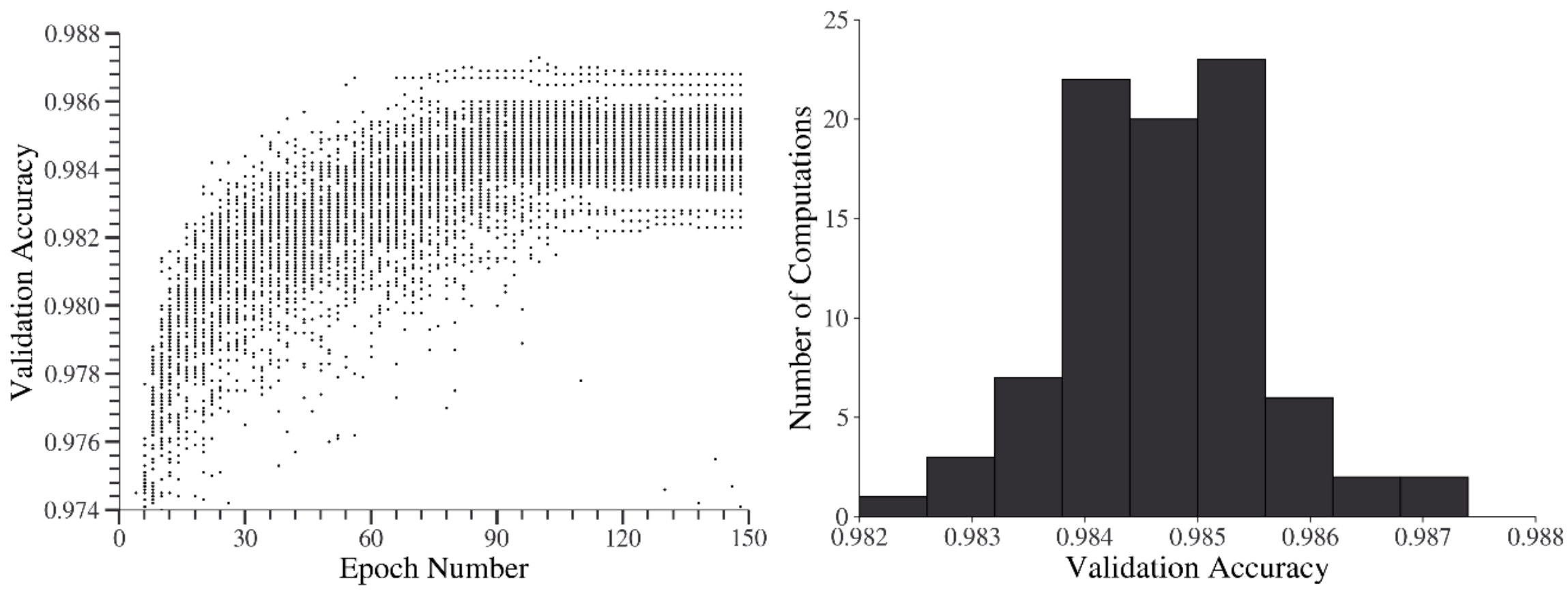

*Figure 6:* As in Figure 1 but for the activation function introduced in this paper with $\gamma = 5$. Note the expanded *x*-axis scale in (b).

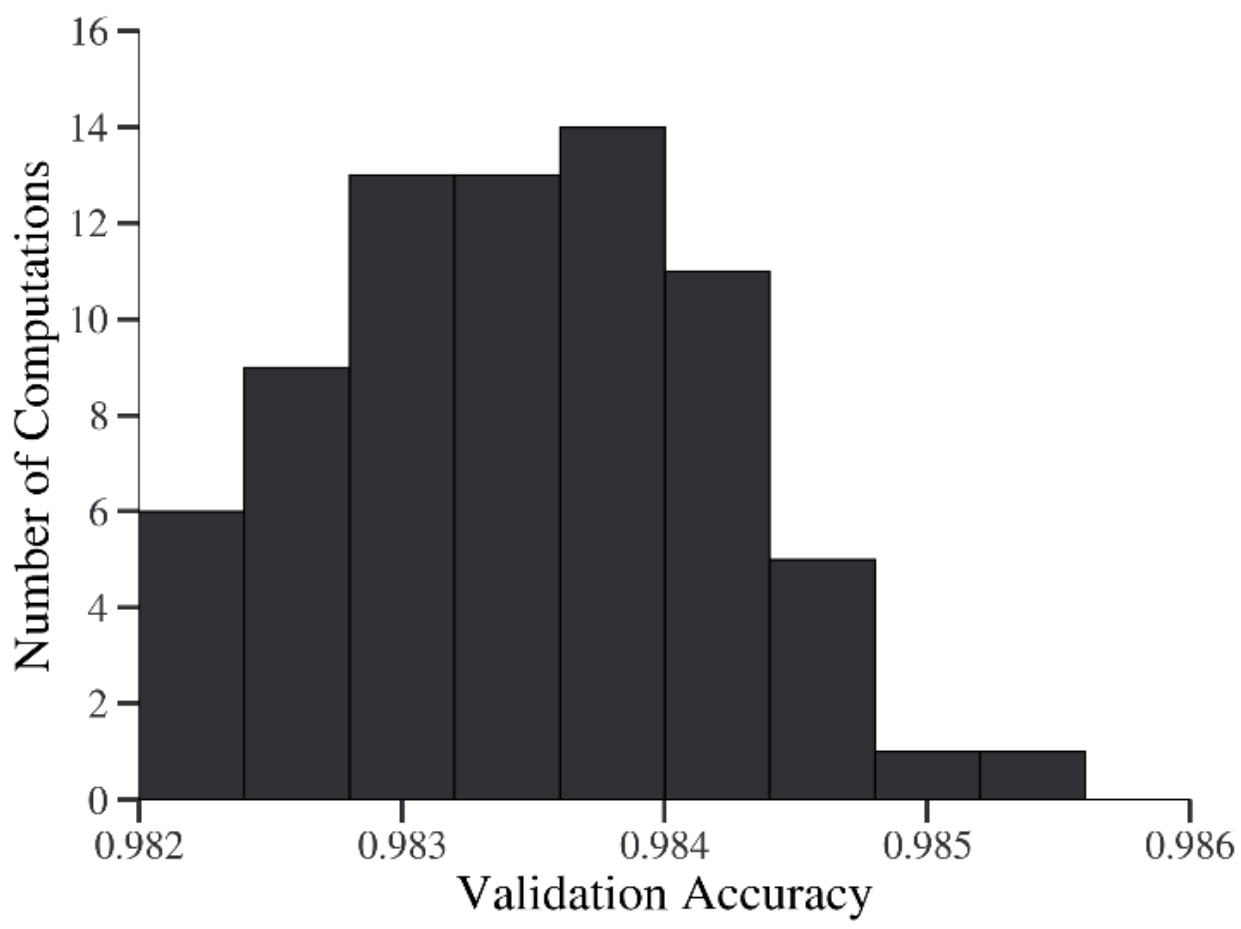

*Figure 7*: The histogram of Figure 6 with $\gamma = 1$ and a cubic term in the activation function for 180 realizations.

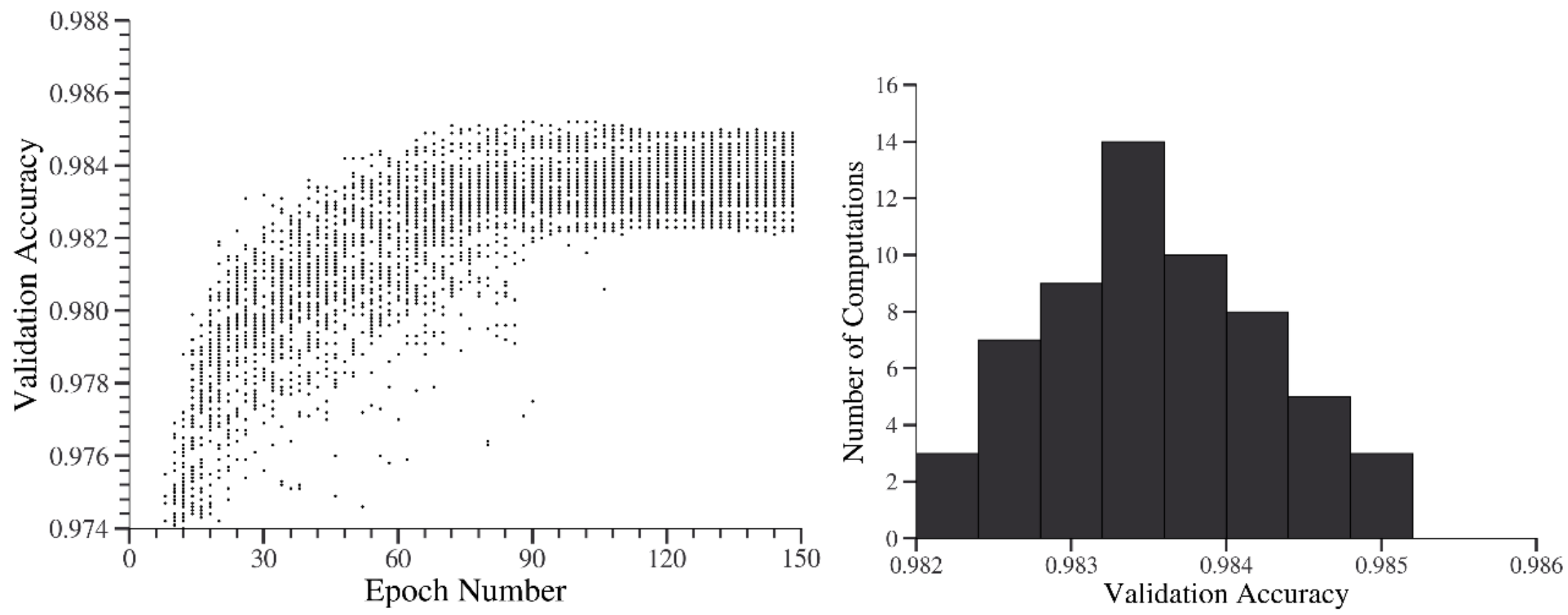

*Figure 8*: As in *Figure 1*, but with a single 512 neuron intermediate layer

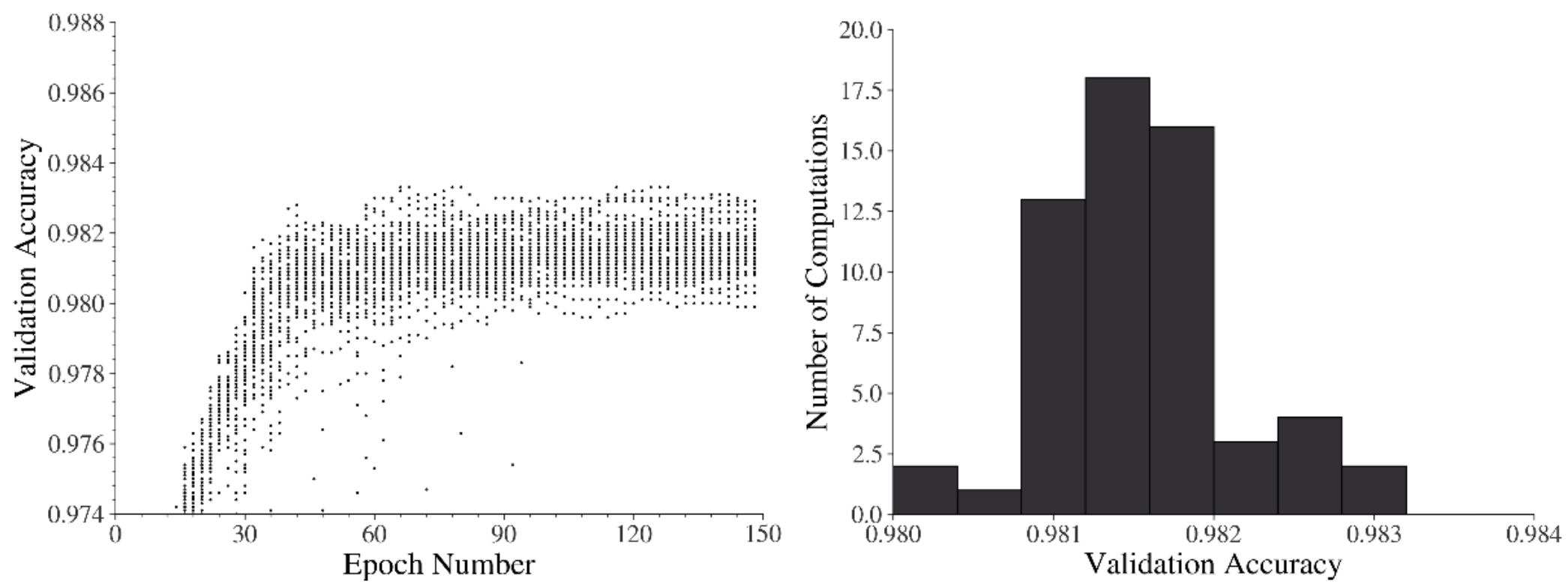

*Figure 9*: As in Figure 1, but for a single layer and an ELU activation function

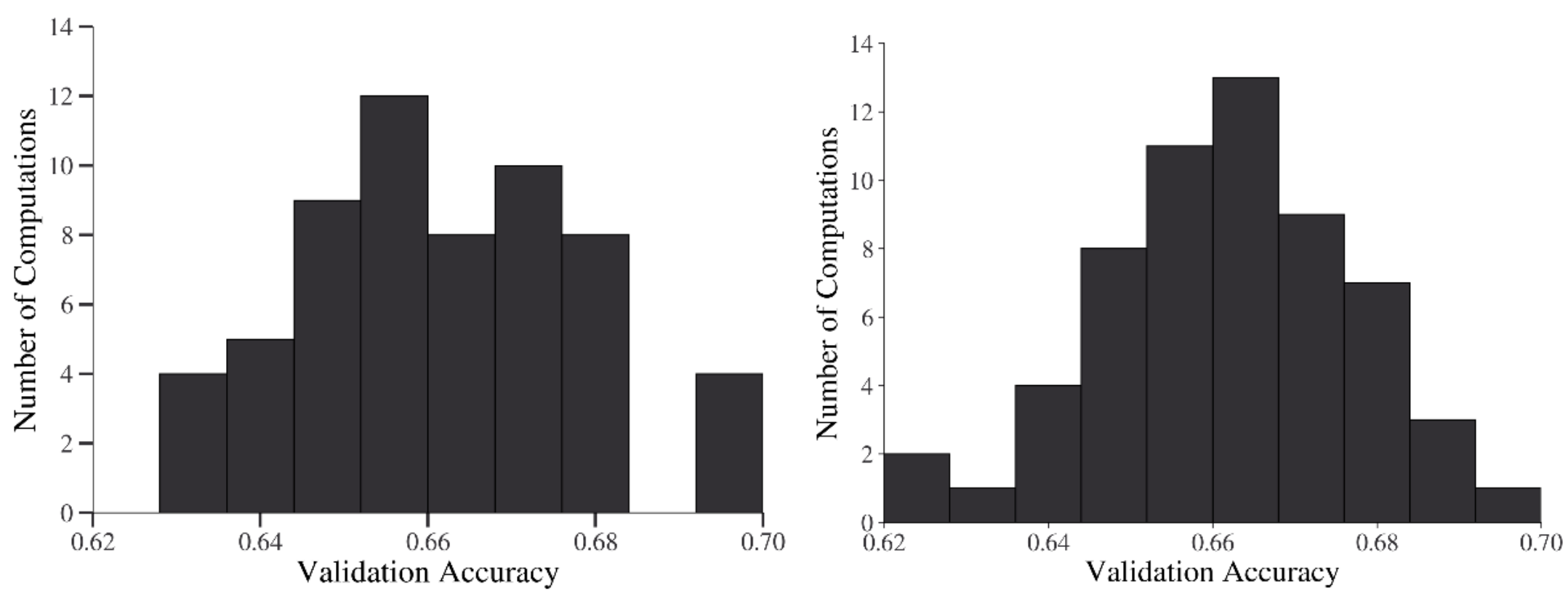

*Figure 10*: The test accuracy of a 4 layer CNN with all ELU layers (left) and a ELU layer followed by ReLU (right) activation functions

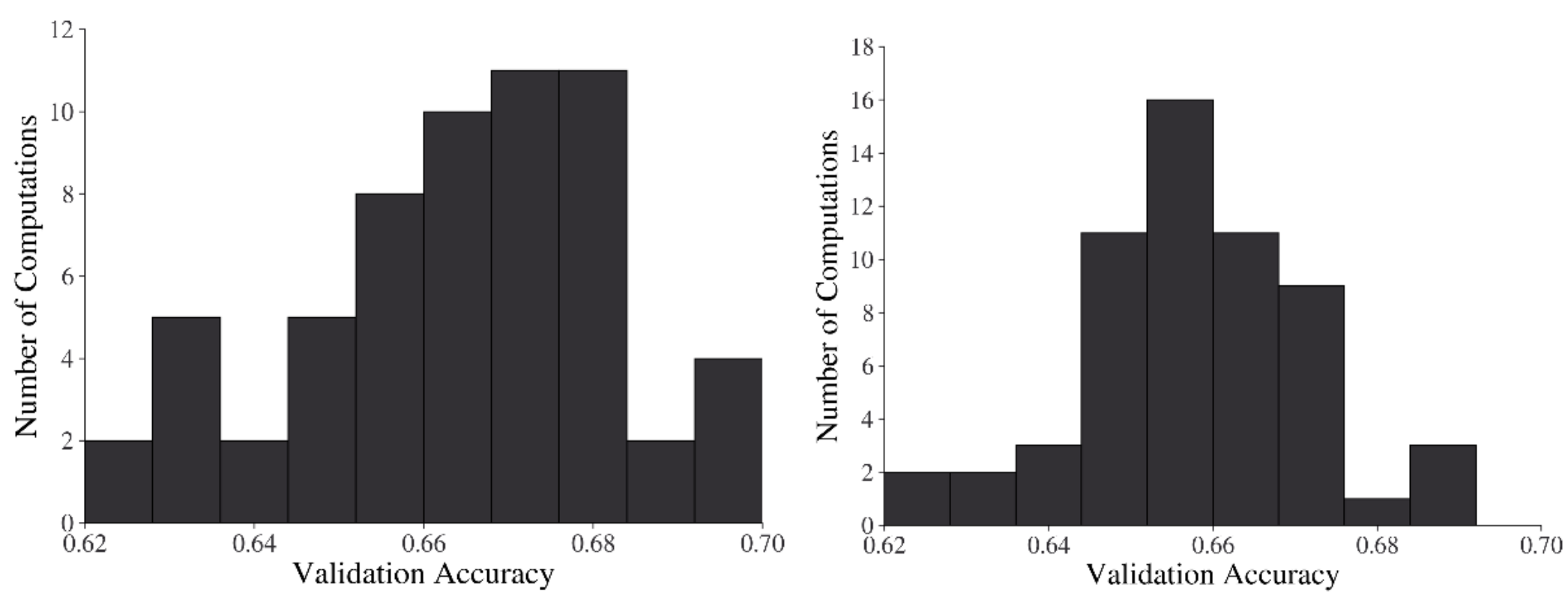

*Figure 11*: As in the right figure of the previous layer but with the ELU function applied to a sum of separately adaptive (left linear and odd quadratic and (right) linear and cubic terms.